\documentclass[journal,10pt]{IEEEtran}
\usepackage[T1]{fontenc}
\usepackage{amsmath,amssymb}
\usepackage{booktabs,multirow,graphicx}
\usepackage{cite}
\usepackage[hidelinks]{hyperref}
\begin{document}
\title{PolarGait: Constructing Complementary Views from a Single Human Sketch for Gait Recognition}
\author{Chao Zhang, Zhuang Zheng, Zhanyong Mei, Huayan Pu, Jun Luo, and Jinglei Zhao%
\thanks{Chao Zhang, Huayan Pu, Jun Luo, and Jinglei Zhao are with the State Key Lab of Mechanical Transmission for Advanced Equipment, Chongqing University, Chongqing 400044, China.}%
\thanks{Zhuang Zheng and Zhanyong Mei are with the College of Computer Science and Cyber Security, Chengdu University of Technology, Chengdu 610059, China.}%
\thanks{Corresponding author: Jinglei Zhao.}}
\maketitle
\begin{abstract}

Vision-based gait recognition has explored diverse representations,
including raw \textbf{\textit{RGB}}, \textbf{\textit{Silhouette}},
\textbf{\textit{Human Parsing}}, \textbf{\textit{Skeleton}}, and
\textbf{\textit{Human Sketch}}. Existing multimodal methods mainly
exploit complementary information by fusing heterogeneous
input representations. However, whether complementary information
can be actively constructed from a single representation before
fusion remains less explored.
Human Sketch provides a suitable basis for this problem because it preserves
both stable body structures and dense local responses within a unified
representation. To this end, we propose \textbf{PolarGait}, which constructs
multiple complementary Sketch views through filtering and spatial
reparameterization and integrates them for gait representation learning.
Specifically, \textbf{Filter-Oriented Sketch Decomposition (FOSD)} applies
Gaussian smoothing to obtain a structure-oriented representation with more
coherent responses, while absolute residual extraction preserves local
discrepancies induced by smoothing. In parallel,
\textbf{Gait-Oriented Polar Transformation (GOPT)} adapts polar
reparameterization to gait sequences through silhouette-guided sketch normalization,
a gait-oriented center, and lower-body-biased non-uniform angular sampling,
providing a spatial view complementary to the original Cartesian
representation. In addition, Silhouette is incorporated as an auxiliary
representation to provide complementary global body-shape information.
Extensive experiments on CCPG, SUSTech1K, CASIA-B, and Gait3D validate the
proposed framework. Under the single-Sketch setting, FOSD+GOPT improves the
original Sketch baseline by \textbf{6.6\%} on CCPG and \textbf{4.6\%} on
CASIA-B. The complete PolarGait achieves Rank-1 performance of
\textbf{87.9\%}, \textbf{95.2\%}, \textbf{96.6\%}, and \textbf{86.4\%} on
the four benchmarks, respectively. These results demonstrate that
complementary gait information need not be restricted to heterogeneous input
modalities, but can also be actively constructed from a single structured
representation and effectively integrated for robust gait recognition.

\end{abstract}


\section{Introduction}

\begin{figure}[t]
    \centering
    \includegraphics[width=1.0\columnwidth,keepaspectratio]
    {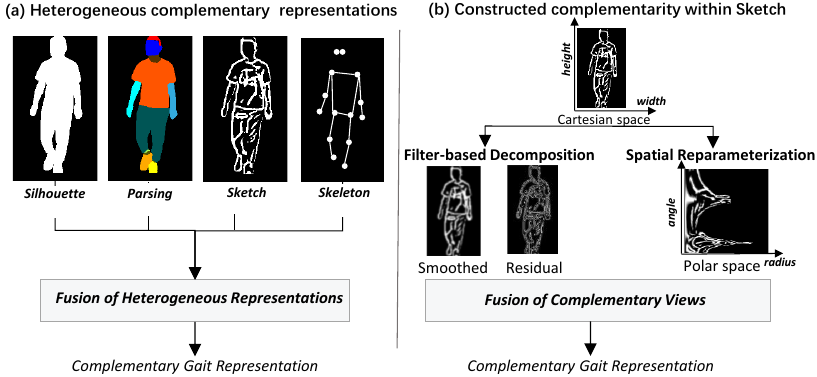}
\caption{Comparison of complementary gait representation learning.
(a) Complementarity across heterogeneous gait representations.
(b) Complementarity constructed within Human Sketch through filtering
and spatial reparameterization.}
\label{fig:motivation}
\end{figure}

Gait recognition identifies individuals from their walking patterns
and has attracted attention due to its non-contact, long-distance,
and non-cooperative capabilities. Unlike face, fingerprint, and iris
recognition, gait can be captured at a distance without subject
cooperation, making it suitable for surveillance and real-world
applications~\cite{bouchrika2017}.

Vision-based gait recognition has achieved substantial progress on both
controlled and in-the-wild benchmarks, partly driven by the continuous
exploration of visual representations derived from RGB videos. Raw
\textbf{\textit{RGB}} images preserve rich appearance and contextual
information but are sensitive to clothing textures, illumination variations,
and background interference~\cite{denoisinggait,biggait}. \textbf{\textit{Silhouette}} suppresses most
appearance information and captures global body shape, but discards internal
structural details~\cite{gaitset,gaitpart,gaitgl}. \textbf{\textit{Human Parsing}} further decomposes the
body into semantic parts, providing richer internal structures while remaining
dependent on upstream parsing quality and label granularity~\cite{parsinggait,gaitparsing}.
\textbf{\textit{Skeleton}} represents the human body through articulated
joints and their spatial relationships, providing compact structural and motion
cues while relying on the accuracy of pose estimation~\cite{gaitgraph,skeletongaitpp}. These representations
encode gait information from different perspectives and exhibit complementary
characteristics, motivating their integration in multimodal gait recognition.

As illustrated in Fig.~\ref{fig:motivation}(a), existing multimodal gait
recognition methods mainly exploit complementary information across
heterogeneous gait representations. BiFusion~\cite{bifusion}, MMGaitFormer~\cite{mmgaitformer}, SkeletonGait++~\cite{skeletongaitpp},
and TriGait~\cite{trigait} combine \textit{Silhouette} and \textit{Skeleton} to jointly
model global body shape and articulated structural information. XGait~\cite{xgait}
integrates \textit{Silhouette} with \textit{Human Parsing} to exploit their
complementary structural properties at different granularities. MultiGait++~\cite{multigaitpp}
and TriFusionGait~\cite{trifusiongait} further extend multimodal gait recognition to trimodal
settings by incorporating \textit{Optical Flow} and \textit{Skeleton},
respectively, thereby introducing additional motion or articulated structural
information into the fusion process. SketchGait~\cite{sketchgait} combines \textit{Human Sketch}
and \textit{Human Parsing} to exploit their complementary structural
characteristics. More recently, GaitFuse~\cite{gaitfuse} further considers cross-modal
semantic discrepancy and modality reliability during multimodal fusion.

These studies demonstrate that heterogeneous gait representations can provide
complementary shape, structural, semantic, and motion information, and that
effective fusion depends not only on combining multiple inputs, but also on
how information from these representations is effectively interacted, aligned, and integrated. Accordingly,
recent multimodal gait methods have increasingly focused on designing more
effective interaction and fusion mechanisms for heterogeneous representations.
However, the complementary information exploited by existing methods is still
primarily provided by heterogeneous input representations, such as the
body-shape information from \textit{Silhouette} and the articulated structural
information from \textit{Skeleton}. It remains less explored whether a single
gait representation can be reorganized into multiple views that emphasize
different information and subsequently provide complementary cues for gait
recognition. This motivates a further question:
\textit{Can useful complementarity also be actively constructed within a
single representation?}

Human Sketch provides a suitable basis for constructing
complementary gait representations. As a dense edge-based
representation, it preserves both external body contours
and internal structural responses without explicit semantic
part labels~\cite{sketchgait}. Compared with
\textit{Silhouette}, \textit{Human Parsing}, and
\textit{Skeleton}, Human Sketch contains a wider range
of structural information, from coherent body contours
to fine-grained local details. However, existing
Sketch-based methods~\cite{sketchgait} primarily exploit
complementarity by combining Sketch with other
representations, leaving the potential for constructing
complementary views within Sketch itself underexplored.

As illustrated in Fig.~\ref{fig:motivation}(b),
we explore two ways to construct complementary
views from a single Sketch: filter-based decomposition
and spatial reparameterization.
First, filter-based decomposition exploits how spatial
filtering affects the information within a Sketch.
Spatial filtering, such as Gaussian smoothing,
preserves recognizable body contours and major
structural lines while suppressing fine-grained
edge responses. This allows us to construct two
complementary views: a smoothed representation
retaining coherent body structures and an absolute
residual capturing response changes induced by filtering.

Second, spatial reparameterization provides another
source of complementarity. Inspired by the Spatial
Polar Transformation (SPT)~\cite{yang2019person}
for person re-identification, we explore whether
Sketch can be reorganized under a different spatial
parameterization. Cartesian and polar spaces describe
the same structures through different spatial
relationships: the former preserves horizontal and
vertical locality, whereas the latter organizes them
by radial and angular relations around the body.
This reorganization changes how the same Sketch
responses are spatially arranged, allowing Cartesian
and polar views to capture complementary structural
patterns without introducing additional visual information.

Building on this analysis, we propose \textbf{PolarGait}
to construct complementary views from Human Sketch.
First, \textbf{Filter-Oriented Sketch Decomposition (FOSD)}
applies Gaussian smoothing and absolute residual extraction
to produce two response views: a smoothed representation
preserving coherent structures and a residual representation
capturing local response variations.
Second, \textbf{Gait-Oriented Polar Transformation (GOPT)}
reorganizes Sketch structures in gait-oriented polar space.
To accommodate frame-wise variations and gait-specific
body characteristics, GOPT employs
\textit{silhouette-guided sketch normalization (SGSN)}
to reduce location and scale variations, a
\textit{gait-oriented center (GOC)} for polar unfolding,
and \textit{lower-body-biased non-uniform angular sampling
(LBB-NAS)} to emphasize lower-limb gait dynamics.
PolarGait then employs three independent branches.
The structure branch combines the smoothed Sketch with
\textit{Silhouette} to incorporate global body-shape cues,
while the residual and polar branches process their
respective representations. Their high-level features
are fused for gait recognition.

Extensive experiments are conducted on CCPG, SUSTech1K, CASIA-B, and
Gait3D. Under the single-Sketch setting, FOSD+GOPT improves the original
Sketch baseline by \textbf{6.6\%} on CCPG and \textbf{4.6\%}
 on CASIA-B, demonstrating the benefit of constructing
complementary representations within Human Sketch. With additional global
body-shape information from Silhouette, PolarGait achieves Rank-1 performance
of \textbf{87.9\%}, \textbf{95.2\%}, \textbf{96.6\%}, and \textbf{86.4\%}
on CCPG, SUSTech1K, CASIA-B, and Gait3D, respectively.

The main contributions of this work are summarized as follows:

\begin{itemize}

\item We demonstrate that complementary gait information
can be constructed within a single Human Sketch,
rather than relying solely on heterogeneous representations.

\item We propose \textbf{PolarGait}, which constructs
complementary Sketch views through \textbf{FOSD}
and \textbf{GOPT}. FOSD reorganizes responses via
Gaussian smoothing and residual extraction, while
GOPT performs gait-oriented polar reparameterization.
Silhouette additionally provides global body-shape cues.

\item Experiments on CASIA-B, SUSTech1K, CCPG, and
Gait3D demonstrate the effectiveness of the constructed
views and strong performance across diverse benchmarks.

\end{itemize}

\section{Related Work}

\subsection{Visual Representations for Gait Recognition}

Vision-based gait recognition has explored diverse visual
representations to capture identity-related information.
\textbf{\textit{Silhouette}} describes global body shape while
suppressing appearance details. GaitSet~\cite{gaitset} employs
set-level aggregation, GaitPart~\cite{gaitpart} models
part-level motion patterns, and GaitGL~\cite{gaitgl} combines
global and local feature learning. TrackletGait~\cite{trackletgait}
further handles unconstrained sequences through tracklet
sampling and wavelet-based downsampling. However, Silhouette
discards internal structures and fine-grained details.
Raw \textbf{\textit{RGB}} retains richer appearance and contextual
information but is sensitive to clothing, illumination, and
background variations. DenoisingGait~\cite{denoisinggait}
suppresses appearance interference, while BigGait~\cite{biggait}
learns gait representations directly from RGB.
BiggerGait~\cite{biggergait} leverages large-scale pretrained
features, and Gait-X~\cite{gaitx} improves gait-related feature
extraction from appearance-rich inputs.

\textbf{\textit{Human Parsing}} provides explicit body-region
semantics. ParsingGait~\cite{parsinggait} and
GaitParsing~\cite{gaitparsing} exploit semantic parts for
fine-grained structural modeling, while
MultiGait++~\cite{multigaitpp} and
TriFusionGait~\cite{trifusiongait} integrate Parsing with
other representations to exploit complementary information.
\textbf{\textit{Skeleton}} represents articulated joints and
their spatiotemporal relationships.
GaitGraph~\cite{gaitgraph} and GaitGraph2~\cite{gaitgraph2}
model joint topology and motion using graph convolutions.
Gait-TR~\cite{gaittr} combines spatial transformation with
temporal convolution, while GaitPT~\cite{gaitpt} employs
hierarchical Transformers. SkeletonGait~\cite{skeletongaitpp}
converts joints into skeleton maps. Despite its reduced
sensitivity to appearance, Skeleton depends on pose
estimation accuracy and retains limited body-shape information.

More recently, \textbf{\textit{Human Sketch}} has emerged as a dense
edge-based representation preserving body contours and
internal structural responses without semantic
labels~\cite{sketchgait}. Unlike other representations,
Sketch captures both coherent body structures and fine
local details within a unified representation. However,
existing Sketch-based methods mainly exploit complementarity
by combining Sketch with additional modalities such as
Human Parsing, leaving complementary view construction
within Sketch itself underexplored.

\subsection{Spatial Reparameterization for Human Representation Learning}

Spatial reparameterization reorganizes visual structures
beyond conventional Cartesian coordinates.
Spatial Transformer Networks (STN)~\cite{stn} introduce
differentiable spatial transformations, while Active
Convolution~\cite{activeconv} and Deformable Convolutional
Networks~\cite{deformconv} learn adaptive sampling locations.
Polar Transformer Networks (PTN) and log-polar
representations~\cite{ptn} further organize visual information
in radial and angular coordinates to model rotation and
scale variations.

For human representation, Spatial Polar Transformation
(SPT)~\cite{yang2019person} transforms human contour sketches
into polar space for discriminative curve learning in person
re-identification. However, generic polar transformation
does not account for frame-wise variations in body location
and scale or the uneven distribution of gait-related motion
cues across body regions. These limitations motivate
gait-oriented spatial reparameterization that provides
a complementary structural view under gait-adapted
polar geometry.

\subsection{Complementary Representation Learning for Gait Recognition}

Information fusion integrates evidence from multiple sources
while addressing uncertainty and heterogeneity~\cite{khaleghi2013fusion}.
In gait recognition, complementary learning primarily
combines heterogeneous representations.
BiFusion~\cite{bifusion} and
SkeletonGait++~\cite{skeletongaitpp} exploit complementary
body-shape and articulated cues from Silhouette and Skeleton,
while MMGaitFormer~\cite{mmgaitformer} performs spatial
and temporal fusion. GaitMA~\cite{gaitma} introduces
cross-modal alignment and interaction, whereas
TriGait~\cite{trigait} adopts a multi-branch architecture.
DAGait~\cite{dagait} further uses Skeleton to guide
Silhouette alignment.

Recent methods explore richer modality combinations.
XGait~\cite{xgait} integrates Silhouette and Human Parsing
through cross-granularity interaction, while
SketchGait~\cite{sketchgait} combines Human Sketch
and Human Parsing to exploit complementary structural cues.
MultiGait++~\cite{multigaitpp} combines Silhouette,
Human Parsing, and Optical Flow to model shared and
modality-specific information, while
TriFusionGait~\cite{trifusiongait} incorporates Skeleton
for additional structural cues.
GaitFuse~\cite{gaitfuse} addresses semantic discrepancy
and modality reliability through hierarchical alignment
and uncertainty-aware fusion.

These methods mainly obtain complementary information
from heterogeneous input representations. In contrast,
we construct complementary response and spatial views
within a single Human Sketch representation, enabling
complementary representation learning before multimodal
fusion.

\begin{figure*}[t]
    \centering
    \includegraphics[
        width=0.85\textwidth,
        height=0.4\textheight,
        keepaspectratio]{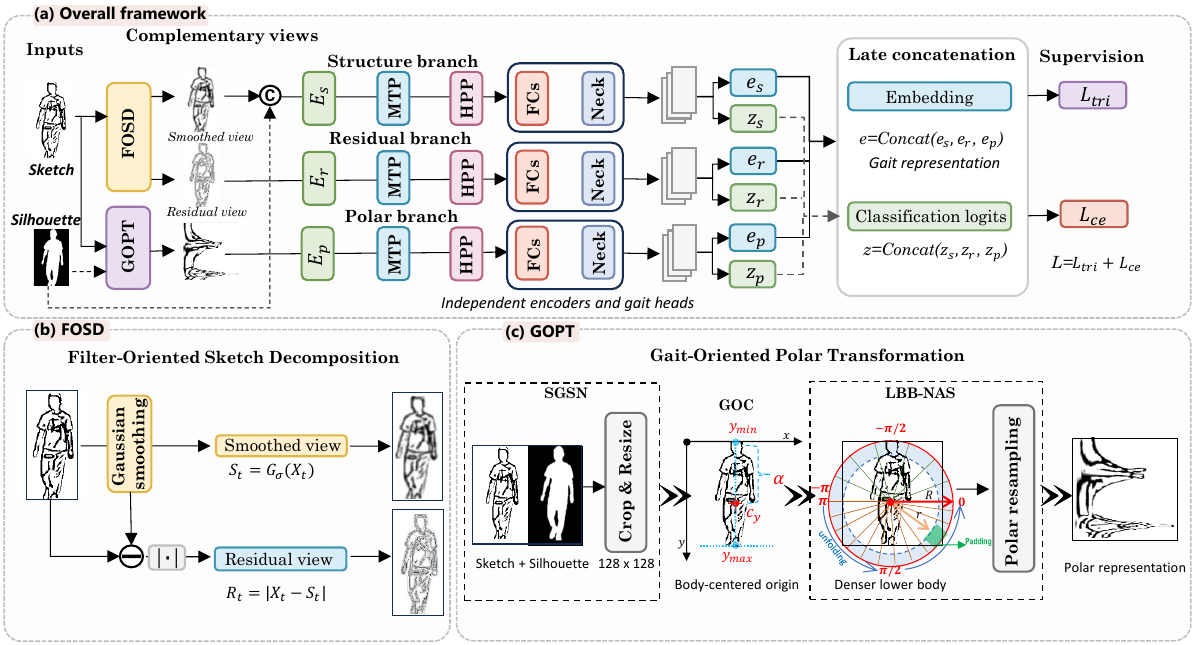}
  \caption{Overview of PolarGait.
(a) Three-branch framework with DeepGaitV2 \cite{deepgaitv2} backbones, following~\cite{gaitset} for MTP, HPP, and FCs, with BNNeck \cite{bnneck}.
(b) Filter-Oriented Sketch Decomposition (FOSD).
(c) Gait-Oriented Polar Transformation (GOPT), where LBB-NAS performs angular unfolding in the order
$\pi \rightarrow \pi/2 \rightarrow 0 \rightarrow -\pi/2 \rightarrow -\pi$.}
    \label{fig:Figures_framework}
\end{figure*}

\section{Method}

\subsection{Overview of PolarGait}

As illustrated in Fig.~\ref{fig:Figures_framework}(a), PolarGait constructs
complementary gait representations centered on Human Sketch through response
reorganization and spatial reparameterization. FOSD produces a smoothed
representation with more coherent structural responses and an absolute residual
representation that captures local response changes induced by smoothing. GOPT
reorganizes the same Sketch in a gait-oriented polar space, providing a
complementary structural description under a different spatial parameterization.

Based on these representations, PolarGait employs three independent branches.
The structure branch combines the smoothed Sketch with the corresponding
Silhouette to incorporate complementary global body shape information. The
residual branch models local response variations induced by smoothing, while
the polar branch learns structural information from the gait-oriented polar
representation.

\subsection{Filter-Oriented Sketch Decomposition}

\paragraph{Motivation}
Human Sketch contains coherent body structures alongside
dense local responses from clothing boundaries, wrinkles,
and textures. These responses behave differently under
spatial filtering: coherent structures remain recognizable
after smoothing, whereas fine-grained patterns undergo
greater local changes. This distinction motivates
constructing complementary views through response
reorganization.

\paragraph{Implementation}
As shown in Fig.~\ref{fig:Figures_framework}(b),
we propose \textbf{Filter-Oriented Sketch Decomposition
(FOSD)} to construct smoothed and absolute residual
views from the same Sketch.

Given a Sketch sequence $\mathbf{X}=\{X_t\}_{t=1}^{T}$,
the smoothed representation is defined as
\begin{equation}
    S_t=\mathcal{G}_{\sigma}(X_t),
\end{equation}
where $\mathcal{G}_{\sigma}(\cdot)$ denotes Gaussian
convolution with standard deviation $\sigma$.
We use a $5\times5$ Gaussian kernel for Sketch inputs
resized to $128\times128$ pixels.

The absolute residual representation is defined as
\begin{equation}
    R_t=\left|X_t-S_t\right|.
\end{equation}
Unlike direct subtraction, the absolute difference
captures both positive and negative response changes
induced by smoothing.

The smoothed view emphasizes coherent structural
responses, while the residual view captures local
response variations. Their independent encoding
enables complementary information to be exploited
for gait recognition.

\subsection{Gait-Oriented Polar Transformation}

\paragraph{Motivation}
Beyond response reorganization, we construct complementary
Sketch views through spatial reparameterization.
Unlike Cartesian coordinates that preserve horizontal
and vertical locality, polar coordinates organize
structures through radial and angular relations~\cite{ptn,yang2019person}.
A fixed convolution window in polar space also corresponds
to a radius-dependent region in Cartesian space, with
tangential extent increasing with radius. This spatial
reorganization enables the same Sketch to be represented
through different structural relationships and spatial scales.

However, generic polar transformation does not account
for frame-wise variations in body location and scale or
the uneven distribution of gait-related cues. In particular,
the upper body often contains denser appearance-related
responses, while the lower limbs exhibit more pronounced
gait dynamics. We therefore propose \textbf{Gait-Oriented
Polar Transformation (GOPT)}, which incorporates
silhouette-guided normalization, a gait-oriented center,
and lower-body-biased non-uniform angular sampling to
adapt polar reparameterization to gait sequences.

\begin{table*}[!t]
    \centering
    \caption{Rank-1 accuracy (\%) on CCPG and SUSTech1K.
CCPG follows the gait recognition protocol, with Mean denoting
the average over four conditions.
OA@R-1 denotes overall Rank-1 accuracy on SUSTech1K.
Best results are in bold; ``--'' denotes unavailable results.}
    \label{tab:comparison_ccpg_sustech}

    \tiny
    \setlength{\tabcolsep}{3.5pt}
    \setlength{\tabcolsep}{2.8pt}
    \renewcommand{\arraystretch}{1.0}

    \resizebox{0.9\textwidth}{!}{%
    \begin{tabular}{@{}llccccc@{\hspace{4pt}}ccccccccc@{}}
        \toprule
        \multirow{2}{*}{Input} &
        \multirow{2}{*}{Method} &
        \multicolumn{5}{c}{CCPG} &
        \multicolumn{9}{c}{SUSTech1K} \\
        \cmidrule(lr){3-7}
        \cmidrule(lr){8-16}
        & &
        CL & UP & DN & BG & Mean &
        NM & BG & CL & CR & UM & UN & OC & NT & OA@R-1 \\
        \midrule

        \multirow{2}{*}{Depth}
        & SimpleView~\cite{simpleview}
        & -- & -- & -- & -- & --
        & 72.3 & 68.8 & 57.2 & 63.3 & 49.2 & 79.7 & 62.5 & 66.5 & 64.8 \\[-1pt]

        & LidarGait~\cite{lidargait}
        & -- & -- & -- & -- & --
        & 91.8 & 88.6 & 74.6 & 89.0 & 67.5 & 94.5 & 80.9 & 90.4 & 86.8 \\
        \midrule

        \multirow{3}{*}{Point Cloud}
        & HMRNet~\cite{hmrnet}
        & -- & -- & -- & -- & --
        & 92.7 & 92.3 & 79.6 & 90.3 & 83.1 & 95.2 & 86.2 & 90.4 & 90.2 \\[-1pt]

        & PointNet++~\cite{pointnetpp}
        & -- & -- & -- & -- & --
        & 82.5 & 78.7 & 58.7 & 76.1 & 74.0 & 85.4 & 75.8 & 74.8 & 77.1 \\[-1pt]

        & LidarGait++~\cite{lidargaitpp}
        & -- & -- & -- & -- & --
        & 94.2 & 93.9 & \textbf{79.7} & 92.4 & 91.5 &
        \textbf{96.6} & 91.9 & \textbf{92.2} & 92.7 \\
        \midrule

        \multirow{6}{*}{Silhouette}
        & GaitSet~\cite{gaitset}
        & 60.2 & 65.2 & 65.1 & 68.5 & 64.8
        & 69.1 & 68.2 & 37.4 & 65.0 & 63.1 & 61.0 & 67.2 & 23.0 & 65.0 \\[-1pt]

        & GaitPart~\cite{gaitpart}
        & 64.3 & 67.8 & 68.6 & 71.7 & 68.1
        & 62.2 & 62.8 & 33.1 & 59.5 & 57.2 & 54.8 & 57.2 & 21.7 & 59.2 \\[-1pt]

        & GaitGL~\cite{gaitgl}
        & -- & -- & -- & -- & --
        & 67.1 & 66.2 & 35.9 & 63.3 & 61.6 & 58.1 & 66.6 & 17.9 & 63.1 \\[-1pt]

        & GaitBase~\cite{gaitbase}
        & 71.6 & 75.0 & 76.8 & 78.6 & 75.5
        & 81.5 & 77.5 & 49.6 & 75.8 & 75.5 & 76.7 & 81.4 & 25.9 & 76.1 \\[-1pt]

        & DeepGaitV2~\cite{deepgaitv2}
        & 78.6 & 84.8 & 80.7 & 89.2 & 83.3
        & 86.5 & 82.8 & 49.2 & 80.4 & 83.3 & 81.9 & 86.0 & 28.0 & 80.9 \\[-1pt]

        & GaitCLIF ~\cite{gaitclif}
        & -- & -- & -- & -- & --
        & 89.2 & 86.6 & 58.0 & 84.0 & 87.7 & 88.0 & 90.7 & 29.9 & 84.7 \\[-1pt]

        & TrackletGait-64 ~\cite{trackletgait}
        & -- & -- & -- & -- & --
        & -- & -- & -- & -- & -- & -- & -- & -- & 86.2  \\
        \midrule

        \multirow{2}{*}{Skeleton}
        & GaitGraph2~\cite{gaitgraph2}
        & 5.0 & 5.3 & 5.8 & 6.2 & 5.6
        & 22.2 & 18.2 & 6.8 & 18.6 & 13.4 & 19.2 & 27.3 & 16.4 & 18.6 \\[-1pt]

        & SkeletonGait~\cite{skeletongaitpp}
        & 40.4 & 48.5 & 53.0 & 61.7 & 50.9
        & 67.9 & 63.5 & 36.5 & 61.6 & 58.1 & 67.2 & 79.1 & 50.1 & 63.0 \\
        \midrule

        Parsing
        & DeepGaitV2~\cite{deepgaitv2}
        & 68.8 & 73.6 & 80.1 & 87.8 & 77.6
        & 93.1 & 87.9 & 40.9 & 88.8 & 87.7 & 88.7 & 93.5 & 47.5 & 87.5 \\
        \midrule

        \multirow{2}{*}{Sketch-T}
        & DeepGaitV2~\cite{deepgaitv2}
        & 61.0 & 67.4 & 75.7 & 83.1 & 71.2
        & 95.2 & 92.1 & 57.4 & 92.3 & 90.8 & 90.6 & 97.9 & 61.2 & 91.3 \\[-1pt]

        & \textbf{FOSD+GOPT}
        & 77.3 & 85.9 & 82.6 & 93.3 & 84.8
        & 96.9 & 95.5 & 71.9 & 94.8 & 94.5 & 93.9 & 98.8 & 69.5 & 94.4 \\

        \multirow{2}{*}{Sketch-M}
        & DeepGaitV2~\cite{deepgaitv2}
        & 70.3 & 78.7 & 81.3 & 89.8 & 80.0
        & 95.5 & 93.6 & 49.7 & 93.3 & 92.7 & 90.7 & 97.5 & 67.4 & 92.1 \\[-1pt]

        & \textbf{FOSD+GOPT}
        & 79.2 & 88.5 & 84.7 & 94.0 & 86.6
        & 96.4 & 95.3 & 63.2 & 94.6 & 94.3 & 92.7 & 98.2 & 70.1 & 93.8 \\
        \midrule

        \multirow{4}{*}{silh.+Skel.}
        & BiFusion~\cite{bifusion}
        & 62.6 & 67.6 & 66.3 & 66.0 & 65.6
        & 69.8 & 62.3 & 45.4 & 60.9 & 54.3 & 63.5 & 77.8 & 33.7 & 62.1 \\[-1pt]

        & SkeletonGait++~\cite{skeletongaitpp}
        & 79.1 & 83.9 & 81.7 & 89.9 & 83.7
        & 85.1 & 82.9 & 46.6 & 81.9 & 80.8 & 82.5 & 86.2 & 47.5 & 81.3 \\[-1pt]

        & PSGait~\cite{psgait}
        & 82.5 & 85.3 & \textbf{88.6} & 89.1 & 86.4
        & 87.9 & 85.2 & 53.9 & 84.5 & 86.3 & 85.3 & 91.4 & 47.5 & 84.6 \\[-1pt]

        & GaitFuse~\cite{gaitfuse}
        & - & - & - & - & -
        & 86.3 & 83.4 & 47.4 & 82.6 & 81.5 & 83.7 & 87.4 & 47.2 & 83.5 \\
        \midrule

        silh.+Pars.
        & XGait~\cite{xgait}
        & 73.9 & 75.5 & 79.8 & 81.2 & 77.6
        & -- & -- & -- & -- & -- & -- & -- & -- & -- \\
        \midrule

        silh.+Pars.+Flow
        & MultiGait++~\cite{multigaitpp}
        & \textbf{83.9} & 89.0 & 86.0 & 91.5 & 87.6
        & 92.0 & 89.4 & 50.4 & 87.6 & 89.7 & 89.1 & 93.4 & 45.1 & 87.4 \\
        \midrule

        Sketch+Pars+silh.
        & SketchGait++ (TEED)~\cite{sketchgait}
        & 82.2 & 87.0 & 86.8 & 94.5 & 87.6
        & \textbf{97.5} & 94.4 & 60.1 & 93.6 & 94.1 & 93.9 & 98.9 & 54.0 & 93.1 \\
        \midrule

        \multirow{2}{*}{Sketch+silh.}
        & \textbf{PolarGait (TEED)}
        & 80.1 & 88.5 & 83.3 & \textbf{94.5} & 86.6
        & 97.4 & \textbf{96.5} & 74.4 &
        \textbf{95.7} & \textbf{95.5} & 94.8 &
        \textbf{98.9} & 74.1 & \textbf{95.2} \\[-1pt]

        & \textbf{PolarGait (MUGE)}
        & 82.1 & \textbf{90.2} & 85.2 & 94.2 & \textbf{87.9}
        & 96.8 & 96.2 & 64.8 & 95.5 & 95.3 & 94.1 & 98.5 & 74.8 & 94.7 \\
        \bottomrule
    \end{tabular}%
    }
\end{table*}

\paragraph{Implementation}
\textbf{Silhouette-Guided Sketch Normalization (SGSN):}
As shown in Fig.~\ref{fig:Figures_framework}(c),
given a Sketch frame $X_t$ and its corresponding silhouette mask $M_t$, we
estimate the vertical foreground range and horizontal mass center from $M_t$.
A square cropping region centered on the body is then applied identically to
both $X_t$ and $M_t$. The cropped Sketch and silhouette are resized to
$128\times128$ pixels, producing the normalized inputs
$\widetilde{X}_t$ and $\widetilde{M}_t$, respectively. This normalization
reduces variations in body location and scale across frames while preserving
thin Sketch structures for subsequent polar resampling.

\textbf{Gait-Oriented Center (GOC):}
The polar center determines how body structures are unfolded and arranged in
polar space. Although SGSN aligns the body spatially,
the fixed center of the $128\times128$ image may not provide
an optimal origin for gait-oriented polar unfolding. We define the GOC according to the normalized
silhouette:
\begin{equation}
\begin{gathered}
c_t^x=\mathcal{M}_x(\widetilde{M}_t), \\
    c_t^y=\widetilde{y}_t^{\min}
    +\alpha\left(
    \widetilde{y}_t^{\max}-\widetilde{y}_t^{\min}
    \right),
\end{gathered}
\end{equation}
where $\mathcal{M}_x(\cdot)$ denotes the horizontal mass center,
$[\widetilde{y}_t^{\min},\widetilde{y}_t^{\max}]$ denotes the vertical
foreground range of $\widetilde{M}_t$, and $\alpha$ controls the normalized
vertical position of the center.

\textbf{Lower-Body-Biased Non-Uniform Angular Sampling (LBB-NAS):}
Given
$\mathbf{c}_t=(c_t^x,c_t^y)$, each polar coordinate $(r,\theta)$ is mapped to
a sampling location in $\widetilde{X}_t$ by
\begin{equation}
    x=c_t^x+r\cos\theta,\qquad
    y=c_t^y+r\sin\theta.
\end{equation}

Before angular sampling, we first determine the radial sampling range.
Given the normalized $128\times128$ Sketch, the initial sampling radius is set to $r=64$ pixels.
Since the gait-oriented center is vertically shifted by $\alpha$ and may deviate from the image center,
we extend the sampling radius by a padding margin of 15 pixels, yielding $R=r+15=79$ pixels,
to ensure sufficient coverage of the entire body.

Uniform angular sampling assigns equal sampling density to different body regions,
despite their unequal contributions to gait dynamics.
To better capture lower-limb motion, LBB-NAS allocates more angular samples to the lower body.
Given $N_{\theta}$ angular samples, we define
\begin{equation}
\begin{gathered}
N_{\mathrm{lower}}
    =\operatorname{round}\!\left(\beta N_{\theta}\right), \\
    N_{\mathrm{upper}}
    =N_{\theta}-N_{\mathrm{lower}},
\end{gathered}
\end{equation}
where $\beta$ controls the proportion of samples allocated to the lower body.
This design provides finer angular resolution for lower-limb structures while retaining
the overall upper-body structure with fewer samples.
The resulting polar representation is resized to $64\times64$ pixels.

\subsection{Complementary Multi-View Representation Learning}

Based on the complementary views constructed by FOSD and
GOPT, PolarGait adopts three independent branches with
DeepGaitV2~\cite{deepgaitv2} backbones.
Each branch employs an independent gait head comprising
Max Temporal Pooling (MTP), Horizontal Pyramid Pooling
(HPP), Separate Fully Connected Layers (FCs)~\cite{gaitset,gaitbase},
and BNNeck~\cite{bnneck}. Let $\mathbf{S}$, $\mathbf{R}$,
$\mathbf{P}$, and $\mathbf{M}$ denote the smoothed Sketch,
residual Sketch, polar representation, and normalized
Silhouette, respectively. The final gait embedding is
obtained through late concatenation:
\begin{equation}
\mathbf{e}_{\mathrm{PolarGait}}
=
\operatorname{Concat}
\left(
\mathcal{F}_{s}([\mathbf{S},\mathbf{M}]_c),
\mathcal{F}_{r}(\mathbf{R}),
\mathcal{F}_{p}(\mathbf{P})
\right),
\end{equation}
where $[\cdot,\cdot]_c$ denotes channel-wise concatenation
and $\mathcal{F}_{s}$, $\mathcal{F}_{r}$, and
$\mathcal{F}_{p}$ represent the independent branch
encoders and gait heads. Classification logits are
likewise concatenated for supervision. This late fusion
integrates complementary structural and spatial information
while preserving branch-specific representations.

\subsection{Training Objective}

PolarGait is optimized using cross-entropy loss and triplet loss~\cite{gaitbase,triplet} to jointly
learn discriminative identity representations across the three branches:
\begin{equation}
\mathcal{L}
=
\mathcal{L}_{\mathrm{ce}}
+
\mathcal{L}_{\mathrm{tri}},
\end{equation}
where $\mathcal{L}_{\mathrm{ce}}$ supervises identity classification and
encourages features from different views to remain discriminative for subject
recognition. The triplet loss $\mathcal{L}_{\mathrm{tri}}$ further constrains
the embedding space by reducing intra-class variation and increasing
inter-class separation.

\begin{table*}[!t]
    \centering
    \caption{Rank-1 accuracy (\%) on CASIA-B and retrieval performance (\%)
    on Gait3D. Sketch (Parsing) denotes a Sketch representation derived
    from Human Parsing. PolarGait (Parsing) uses this representation
    without FOSD.}
    \label{tab:comparison_casiab_gait3d}

    \tiny
    \setlength{\tabcolsep}{2.8pt}
    \renewcommand{\arraystretch}{1.0}

    \resizebox{0.6\textwidth}{!}{%
    \begin{tabular}{@{}llcccc@{\hspace{8pt}}cccc@{}}
        \toprule
        \multirow{2}{*}{Input} &
        \multirow{2}{*}{Method} &
        \multicolumn{4}{c}{CASIA-B} &
        \multicolumn{4}{c}{Gait3D} \\
        \cmidrule(lr){3-6}
        \cmidrule(lr){7-10}
        & & NM & BG & CL & Mean &
        Rank-1 & Rank-5 & mAP & mINP \\
        \midrule

        \multirow{2}{*}{Skeleton}
        & GaitGraph2~\cite{gaitgraph2}
        & 82.0 & 73.2 & 63.6 & 72.9
        & -- & -- & -- & -- \\[-1pt]

        & GaitGraph~\cite{gaitgraph}
        & -- & -- & -- & --
        & 6.3 & 16.2 & 5.2 & 2.4 \\
        \midrule

        \multirow{7}{*}{Silhouette}
        & GaitSet~\cite{gaitset}
        & 96.1 & 90.8 & 70.3 & 85.7
        & 36.7 & 58.3 & 30.0 & 17.3 \\[-1pt]

        & GaitPart~\cite{gaitpart}
        & 96.2 & 91.5 & 78.7 & 88.8
        & 28.2 & 47.6 & 21.6 & 12.4 \\[-1pt]

        & GaitGL~\cite{gaitgl}
        & 97.4 & 94.5 & 83.6 & 91.8
        & 29.7 & 48.5 & 22.3 & 13.3 \\[-1pt]

        & GaitBase~\cite{gaitbase}
        & 97.6 & 94.0 & 77.4 & 89.7
        & 64.6 & -- & -- & -- \\[-1pt]

        & DeepGaitV2~\cite{deepgaitv2}
        & -- & -- & -- & 89.6
        & 74.4 & 88.0 & 65.8 & 39.2 \\

        & GaitCLIF~\cite{gaitclif}
        & -- & - & -- & --
        & 76.5 & 88.5 & 67.9 & -- \\[-1pt]

        & TrackletGait-64~\cite{trackletgait}
        & -- & - & -- & --
        & 77.8 & 89.0 & 70.2 & -- \\[-1pt]
        \midrule

        \multirow{2}{*}{Parsing}
        & GaitBase~\cite{gaitbase}
        & -- & -- & -- & --
        & 73.8 & 88.6 & 66.4 & -- \\[-1pt]

        & ParsingGait~\cite{parsinggait}
        & -- & -- & -- & --
        & 76.2 & 89.1 & 68.2 & 41.3 \\
        \midrule

        \multirow{2}{*}{Sketch (MUGE)}
        & DeepGaitV2~\cite{deepgaitv2}
        & 99.5 & 98.2 & 76.5 & 91.4
        & -- & -- & -- & -- \\[-1pt]

        & FOSD+GOPT
        & 99.8 & 99.0 & 89.2 & 96.0
        & -- & -- & -- & -- \\
        \midrule

        \multirow{1}{*}{Sketch (Parsing)}
        & DeepGaitV2~\cite{deepgaitv2}
        & -- & -- & -- & --
        & 82.7 & 92.4 & 76.1 & 59.0 \\
        \midrule

        \multirow{3}{*}{Silh.+Skel.}
        & BiFusion~\cite{bifusion}
        & 98.7 & 96.0 & \textbf{92.0} & 95.6
        & -- & -- & -- & -- \\[-1pt]

        & SkeletonGait++~\cite{skeletongaitpp}
        & -- & -- & -- & --
        & 77.6 & 89.4 & 70.3 & 42.6 \\

        & GaitFuse~\cite{gaitfuse}
        & -- & -- & -- & --
        & 81.2 & 90.8 & 71.2 & 43.5 \\
        \midrule

        Silh.+Pars.
        & XGait~\cite{xgait}
        & -- & -- & -- & --
        & 80.5 & 91.9 & 73.3 & 55.4 \\[-1pt]

        Silh.+Pars.+Flow
        & MultiGait++~\cite{multigaitpp}
        & -- & -- & -- & --
        & 85.4 & \textbf{94.9} & 80.5 & \textbf{65.2} \\
        \midrule

        \multirow{2}{*}{Sketch+Silh.}
        & PolarGait (MUGE)
        & \textbf{99.9} & \textbf{99.2} & 90.6 & \textbf{96.6}
        & -- & -- & -- & -- \\[-1pt]

        & PolarGait (Parsing)
        & -- & -- & -- & --
        & \textbf{86.4} & 93.7 & \textbf{80.7} & 64.4 \\

        \bottomrule
    \end{tabular}%
    }
\end{table*}

 \section{Experiments}

\subsection{Experimental Setup}

\subsubsection{Datasets and Evaluation Protocols}

We evaluate PolarGait on four gait recognition benchmarks:
CCPG, SUSTech1K, CASIA-B, and Gait3D, covering controlled
indoor, challenging outdoor, and in-the-wild scenarios.

\textbf{CCPG}~\cite{ccpg} contains 200 subjects under seven
clothing conditions and up to ten viewpoints. Following
the official protocol, 100 subjects are used for training
and 100 for testing. We report Rank-1 accuracy under
full clothing change (CL), upper-body change (UP),
lower-body change (DN), and backpack (BG), along with
their mean rank-1 accuracy.

\textbf{SUSTech1K}~\cite{lidargait} contains 1,050 subjects
and 25,239 sequences across eight conditions: normal
walking (NM), bag carrying (BG), clothing variation (CL),
carrying objects (CR), umbrella (UM), uniform (UN),
occlusion (OC), and nighttime (NT). Following the official
split, 250 subjects are used for training and 800 for testing.
We report condition-wise Rank-1 accuracy and overall
Rank-1 accuracy (OA@R-1).

\textbf{CASIA-B}~\cite{casiab} contains 124 subjects
recorded from 11 viewpoints, each with six normal
walking (NM), two bag-carrying (BG), and two
clothing-changing (CL) sequences. We use 74 subjects
for training and 50 for testing. Following the standard
protocol, NM\#1--4 form the gallery, while NM\#5--6,
BG\#1--2, and CL\#1--2 serve as probes. We report
average cross-view Rank-1 accuracy, excluding
identical-view matching.

\textbf{Gait3D}~\cite{gait3d} contains 4,000 subjects
and 25,309 sequences captured by 39 cameras under
unconstrained conditions. Following the official protocol,
3,000 subjects (18,940 sequences) are used for training
and 1,000 for testing. Evaluation metrics include Rank-1,
Rank-5, mean Average Precision (mAP), and mean
Inverse Negative Penalty (mINP).

\subsubsection{Sketch Generation and Preprocessing}

For Sketch generation, we adopt TEED~\cite{teed} following
SketchGait~\cite{sketchgait} and additionally employ
MUGE~\cite{muge} to evaluate generalization across
different edge extractors.

For SUSTech1K and CASIA-B, the provided silhouettes are
not aligned with the original RGB frames. We therefore
use SAM3~\cite{sam3} to generate aligned foreground
masks and remove background pixels before Sketch extraction.
Due to the low resolution of CASIA-B,
we employ Real-ESRGAN~\cite{realesrgan}
to upscale RGB frames by a factor of four
before Sketch extraction.

Since Gait3D does not provide original RGB frames, we
follow SketchGait~\cite{sketchgait} to derive coarse Sketches
from Human Parsing using morphological dilation and
erosion, termed \textit{parsing-to-edge}.

\begin{table*}[t]
    \centering
    \caption{Ablation study of the main components on CCPG in terms of
    Rank-1 accuracy (\%). Sketch $\times 2$ denotes a dual-branch
    baseline with identical original Sketch inputs.
    \emph{Silh. Fusion} indicates the FOSD branch with which
    the Silhouette is concatenated.}
    \label{tab:main_component_ablation}

    \scriptsize
    \setlength{\tabcolsep}{2.8pt}
    \renewcommand{\arraystretch}{1.0}

    \begin{tabular}{@{}ccccc@{\hspace{10pt}}cccccc@{}}
        \toprule
        \multirow{2}{*}{Sketch} &
        \multicolumn{2}{c}{FOSD} &
        \multirow{2}{*}{GOPT} &
        \multirow{2}{*}{Silh. Fusion} &
        \multicolumn{6}{c}{CCPG} \\
        \cmidrule(lr){2-3}
        \cmidrule(lr){6-11}
        & Smoothed & Residual & & &
        CL & UP & DN & BG & Mean & $\Delta$ Mean \\
        \midrule

        $\checkmark$ & & & & -- &
        70.3 & 78.7 & 81.3 & 89.8 & 80.0 & -- \\

        $\checkmark\times 2$ & & & & -- &
        73.6 & 82.4 & 83.2 & 92.0 & 82.8 & +2.8 \\

        & $\checkmark$ & & & -- &
        64.4 & 74.8 & 78.6 & 89.0 & 76.7 & -3.3 \\

        & & $\checkmark$ & & -- &
        65.4 & 76.1 & 78.2 & 87.2 & 76.7 & -3.3 \\

        & $\checkmark$ & $\checkmark$ & & -- &
        76.1 & 84.1 & 84.1 & 92.6 & 84.2 & +4.2 \\

        & & & $\checkmark$ & -- &
        67.4 & 78.7 & 66.6 & 82.8 & 73.9 & -6.1 \\

        $\checkmark$ & & & $\checkmark$ & -- &
        78.2 & 87.4 & 81.6 & 92.6 & 85.0 & +5.0 \\

        & $\checkmark$ & $\checkmark$ & $\checkmark$ & -- &
        79.2 & 88.5 & 84.7 & 94.0 & 86.6 & +6.6 \\

        & $\checkmark$ & $\checkmark$ & $\checkmark$ & Residual &
        81.8 & 89.9 & 84.9 & \textbf{94.3} & 87.7 & +7.7 \\

        & $\checkmark$ & $\checkmark$ & $\checkmark$ & Smoothed &
        \textbf{82.1} & \textbf{90.2} & \textbf{85.2}
        & 94.2 & \textbf{87.9} & \textbf{+7.9} \\

        \bottomrule
    \end{tabular}
\end{table*}

\subsubsection{Implementation Details}

All models are trained using SGD with momentum 0.9,
weight decay $5\times10^{-4}$, and an initial learning
rate of 0.1, decayed by a factor of 0.1 at predefined
milestones. For CASIA-B, CCPG, and Gait3D, training
runs for 120K iterations with learning rate decay at
40K, 80K, and 100K, using a batch configuration of
$8\times8\times30$. For SUSTech1K, training runs for
50K iterations with decay at 20K, 30K, and 40K,
using $8\times8\times10$. The batch configuration
denotes identities $\times$ sequences per identity
$\times$ frames per sequence.

\subsection{Comparison with State-of-the-Art Methods}

\subsubsection{Single-Sketch Performance}

We evaluate FOSD+GOPT using only Sketch as the encoded
representation, with Silhouette employed solely for geometric
normalization in GOPT. This setting allows us to evaluate the effectiveness of complementary view construction within Human Sketch without additional Silhouette feature fusion.

As shown in Tables~\ref{tab:comparison_ccpg_sustech} and
\ref{tab:comparison_casiab_gait3d}, FOSD+GOPT consistently
improves the original Sketch baseline under the same
DeepGaitV2~\cite{deepgaitv2} backbone. On SUSTech1K,
OA@R-1 increases from 91.3\% to 94.4\% for TEED and
from 92.1\% to 93.8\% for MUGE, with TEED gaining
\textbf{14.5\%} and \textbf{8.3\%}  under CL and NT,
respectively. On CCPG, the mean Rank-1 accuracy improves
by \textbf{13.6\%} and \textbf{6.6\% } for TEED and MUGE,
respectively. On CASIA-B, MUGE improves from 91.4\%
to 96.0\%.

Compared with the strongest existing single-modal methods,
FOSD+GOPT achieves gains of 1.7\%, 3.3\%, and 4.2\%  on SUSTech1K, CCPG, and CASIA-B, respectively.
These results demonstrate the effectiveness of constructing
complementary response and spatial views from a single
Sketch representation.

\subsubsection{Overall Performance of PolarGait}

We evaluate the complete PolarGait by incorporating
Silhouette into the smoothed structure branch to
complement the constructed Sketch views. As shown in
Tables~\ref{tab:comparison_ccpg_sustech} and
\ref{tab:comparison_casiab_gait3d}, this integration
further improves performance over FOSD+GOPT.

On SUSTech1K, PolarGait achieves OA@R-1 of 95.2\%
and 94.7\% with TEED and MUGE, respectively, improving
FOSD+GOPT by 0.8\% and 0.9\%. The TEED
variant also exceeds the previous best result by 2.1\%. On CCPG, PolarGait (MUGE) achieves
87.9\% mean Rank-1 accuracy, surpassing FOSD+GOPT
by 1.3\% and MultiGait++ by 0.3\%. On CASIA-B, PolarGait reaches
96.6\% mean Rank-1 accuracy, a further 0.6\% gain.

On Gait3D, where RGB frames are unavailable, we employ
\textit{parsing-to-edge} Sketches and construct a complementary
polar view using GOPT. Combined with Silhouette,
PolarGait achieves 86.4\% Rank-1, 93.7\% Rank-5,
80.7\% mAP, and 64.4\% mINP, attaining the highest
Rank-1 accuracy among the compared methods.
These results demonstrate the effectiveness of
integrating constructed Sketch views with global
body-shape information across diverse gait benchmarks.

\subsection{Analysis of Constructed Complementarity}

We investigate the effectiveness of the complementary views
constructed by FOSD and GOPT, as well as the contribution of
additional Silhouette information. Unless otherwise specified,
all ablation experiments are conducted on CCPG using MUGE
Sketches with the same backbone and training settings.
The corresponding results are reported in
Table~\ref{tab:main_component_ablation}.

\begin{figure*}[ht]
    \centering
    \includegraphics[width=0.7\textwidth]{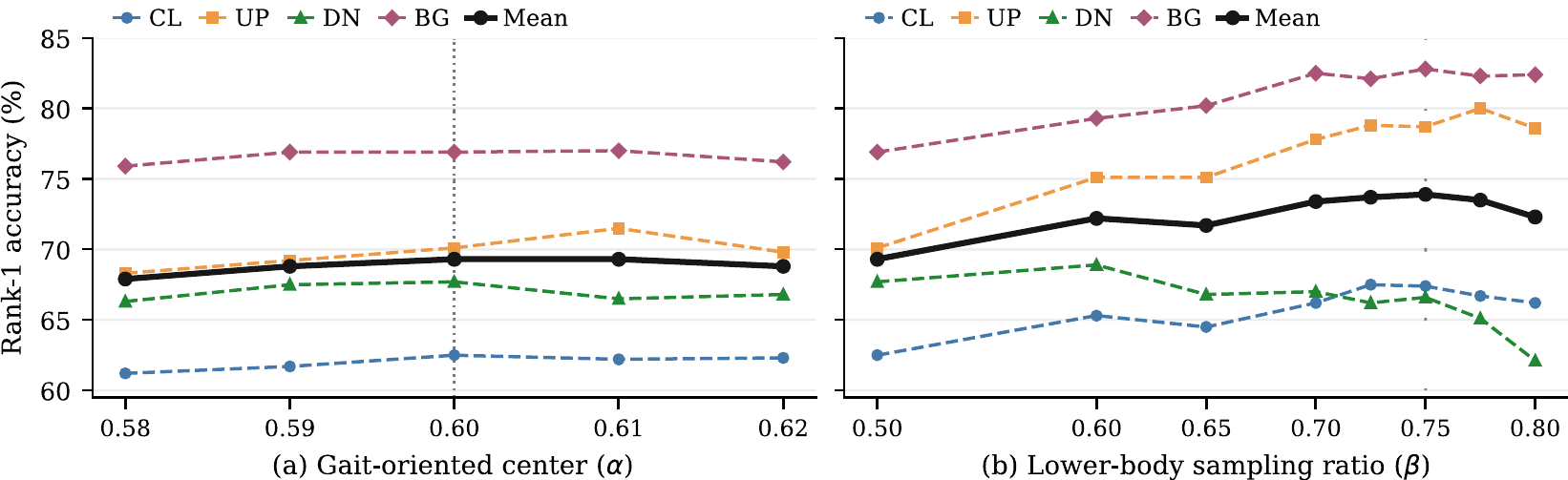}
    \caption{Parameter sensitivity analysis on CCPG.
    (a) Ablation study of the gait-oriented center parameter $\alpha$.
    (b) Ablation study of the lower-body sampling ratio $\beta$.}
    \label{fig:parameter_ablation}
\end{figure*}

\subsubsection{Analysis of FOSD}

We first evaluate the effectiveness of FOSD in constructing
complementary Sketch views. As shown in
Table~\ref{tab:main_component_ablation}, FOSD improves
the mean Rank-1 accuracy from 80.0\% to 84.2\% on CCPG,
achieving a gain of 4.2\%. In contrast,
the smoothed and residual views each achieve only 76.7\%
when used independently. Their superior combined performance
suggests that the two views capture complementary
discriminative information rather than simply providing
stronger individual representations.

To distinguish the benefit of view construction from
increased model capacity, we further introduce a
Sketch $\times 2$ baseline, where two independent branches
process identical original Sketch inputs. Under the same
network configuration and training settings, this baseline
achieves 82.8\%, improving the original Sketch by 2.8\%. FOSD further increases the accuracy
to 84.2\%, outperforming Sketch $\times 2$ by 1.4\%. This result indicates that the gain
of FOSD cannot be attributed solely to increased model
capacity, providing further evidence for the effectiveness
of constructing complementary response views.

\begin{table}[t]
    \centering
    \caption{Comparison of polar center strategies on CCPG
    under uniform angular sampling. Rank-1 accuracy (\%).}
    \label{tab:polar_center_comparison}
    \setlength{\tabcolsep}{3.5pt}
    \renewcommand{\arraystretch}{1.05}
    \resizebox{0.7\columnwidth}{!}{%
    \begin{tabular}{lccccc}
        \toprule
        Center & CL & UP & DN & BG & Mean \\
        \midrule
        Image center
        & 60.7 & 65.1 & 68.6 & 74.6 & 67.3 \\

        BBox center
        & 60.7 & 66.7 & 68.0 & 75.6 & 67.7 \\

        Silhouette centroid
        & 61.6 & 67.0 & \textbf{69.4}
        & \textbf{77.0} & 68.7 \\

        \textbf{GOC ($\alpha=0.60$)}
        & \textbf{62.5} & \textbf{70.1}
        & 67.7 & 76.9 & \textbf{69.3} \\
        \bottomrule
    \end{tabular}%
    }
\end{table}

\subsubsection{Analysis of GOPT}

We analyze the gait-oriented center (GOC) and lower-body-biased
non-uniform angular sampling (LBB-NAS) on CCPG, with other
settings fixed.
For GOC, we first compare four polar center strategies:
image center, bounding-box center, Silhouette centroid,
and the proposed gait-oriented center. Unlike the image-centered SPT~\cite{yang2019person},
GOC determines the polar origin based on body geometry
to better accommodate gait-specific structural characteristics. As shown in
Table~\ref{tab:polar_center_comparison}, GOC achieves
the highest mean Rank-1 accuracy of 69.3\%, exceeding
the image, bounding-box, and centroid alternatives by
2.0\%, 1.6\%, and 0.6\%, respectively.
The gains are most evident under CL and UP.

We further vary $\alpha$ from 0.58 to 0.62 under uniform
angular sampling. Fig.~\ref{fig:parameter_ablation}(a)
shows that $\alpha=0.60$ and $0.61$ achieve the highest
mean accuracy, with $\alpha=0.60$ performing better
under CL and DN. We therefore adopt $\alpha=0.60$.

For LBB-NAS, we fix $\alpha=0.60$ and vary the proportion
$\beta$ of angular samples assigned to the lower body.
As shown in Fig.~\ref{fig:parameter_ablation}(b),
increasing $\beta$ from 0.50 to 0.75 improves the mean
Rank-1 accuracy from 69.3\% to 73.9\%, with gains of
4.9\%, 8.6\%, and 5.9\% under CL, UP, and BG,
respectively. Further increasing $\beta$ reduces
performance, particularly under DN. This suggests that
denser lower-body sampling benefits gait representation,
while excessive allocation weakens the balance between
upper- and lower-body information. We therefore set
$\beta=0.75$ by default.

\begin{table}[!t]
    \centering
    \caption{Ablation results with different Sketch extractors
    on SUSTech1K.}
    \label{tab:granularity_ablation}

    \tiny
    \setlength{\tabcolsep}{2.5pt}
    \renewcommand{\arraystretch}{1.05}

    \resizebox{0.9\columnwidth}{!}{%
    \begin{tabular}{@{}lcccccccc@{}}
        \toprule
        \multirow{2}{*}{Configuration}
        & \multicolumn{4}{c}{TEED}
        & \multicolumn{4}{c}{MUGE} \\
        \cmidrule(lr){2-5}
        \cmidrule(lr){6-9}
        & NM & CL & NT & OA@R-1
        & NM & CL & NT & OA@R-1 \\
        \midrule

        Baseline
        & 95.2 & 57.4 & 61.2 & 91.3
        & 95.5 & 49.7 & 67.4 & 92.1 \\

        +FOSD
        & 95.7 & 59.5 & 62.1 & 91.9
        & 95.5 & 49.5 & 62.1 & 92.3 \\

        +FOSD+Silh.
        & \textbf{96.6} & \textbf{65.6}
        & \textbf{70.8} & \textbf{93.8}
        & \textbf{96.0} & \textbf{54.0}
        & \textbf{73.2} & \textbf{93.6} \\

        \bottomrule
    \end{tabular}%
    }
\end{table}

\begin{figure}[ht]
    \centering
    \includegraphics[
        width=0.85\columnwidth,
        height=0.4\textheight,
        keepaspectratio
    ]{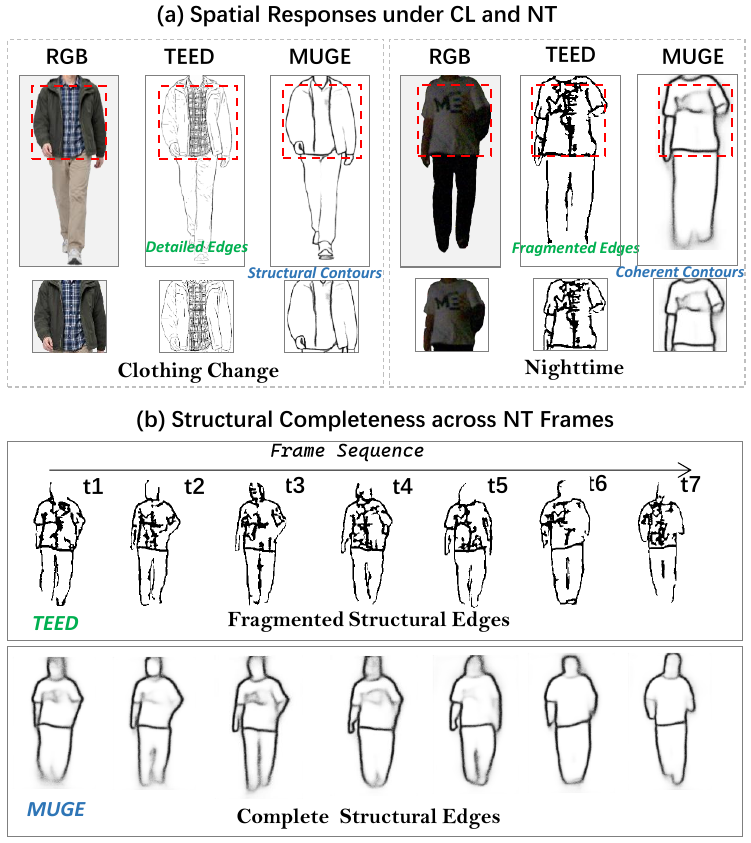}
    \caption{Qualitative comparison of TEED and MUGE on SUSTech1K under clothing change and nighttime conditions.}
    \label{fig:representation_statistics}
\end{figure}
\subsection{Generalization and Robustness}
\begin{figure}[t]
    \centering
    \includegraphics[
        width=1\columnwidth,
        height=0.4\textheight,
        keepaspectratio
    ]{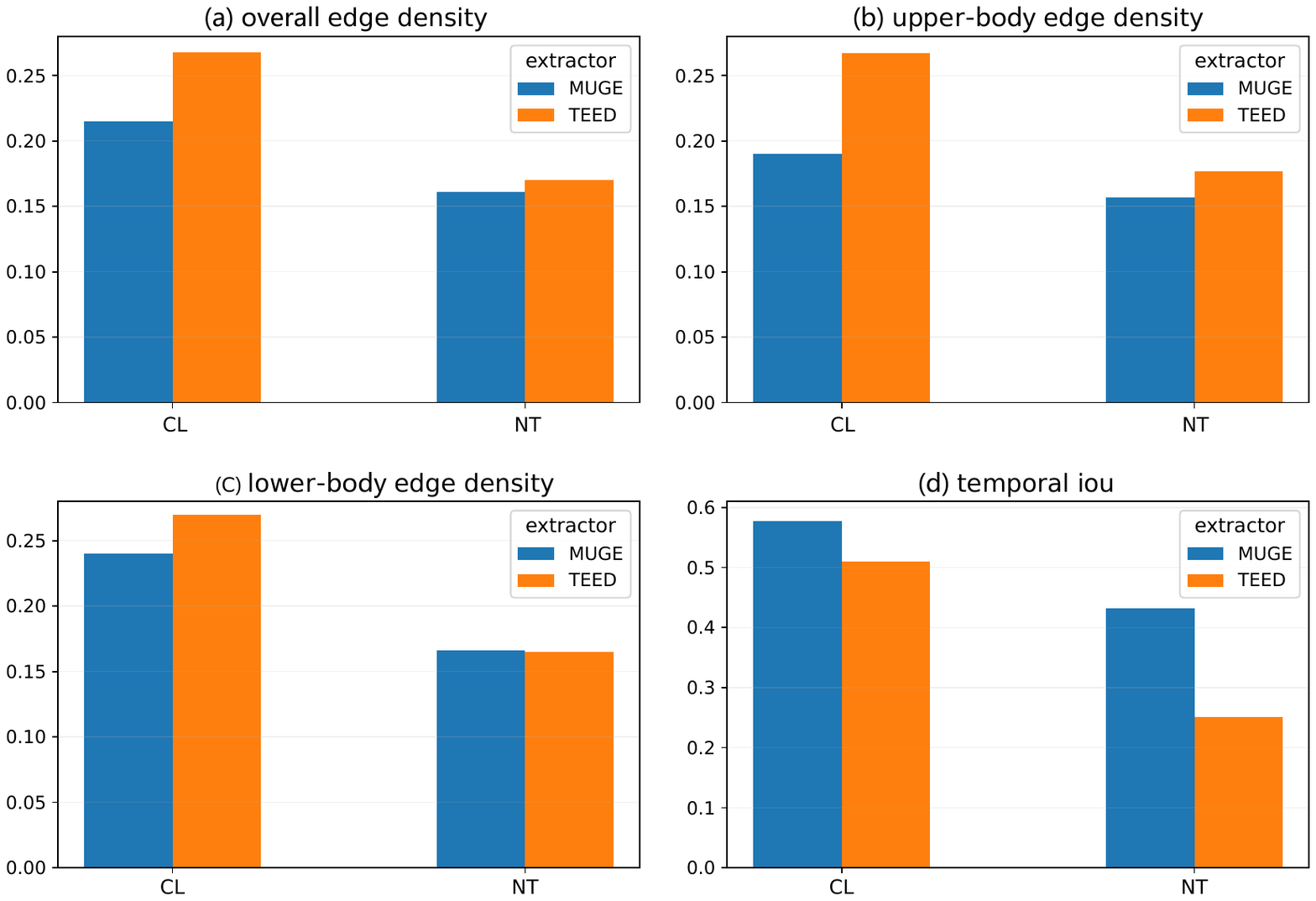}
    \caption{Quantitative comparison of TEED and MUGE under clothing change (CL) and nighttime (NT) conditions.
(a) Overall edge density.
(b) Upper-body edge density.
(c) Lower-body edge density.
(d) Temporal intersection over union (IoU).}
    \label{fig:edge_density}
\end{figure}

\subsubsection{Robustness Across Sketch Extractors}

We investigate how different Sketch extractors affect
the performance of PolarGait. As shown in
Table~\ref{tab:granularity_ablation}, TEED outperforms
MUGE under clothing changes (57.4\% vs. 49.7\%),
whereas MUGE performs better at nighttime
(67.4\% vs. 61.2\%). Although MUGE achieves higher
baseline OA@R-1 (92.1\% vs. 91.3\%), TEED gains
more from FOSD (+0.6\% vs. +0.2\%)
and reaches 93.8\% with additional Silhouette,
slightly exceeding MUGE (93.6\%).

As illustrated in Fig.~\ref{fig:representation_statistics},
TEED preserves richer internal responses under clothing
changes but produces more fragmented edges at nighttime,
whereas MUGE maintains cleaner and more coherent contours.
To quantify these differences, we measure edge density
within body region $R$ and temporal intersection over
union (T-IoU) between consecutive Sketch frames:
\begin{equation}
D(R)=
\frac{\sum_t |E_t \cap M_t^{R}|}
     {\sum_t |M_t^{R}|},
\end{equation}
\begin{equation}
\mathrm{T\text{-}IoU}
=
\frac{1}{T-1}
\sum_{t=1}^{T-1}
\frac{|E_t \cap E_{t+1}|}
     {|E_t \cup E_{t+1}|},
\end{equation}
where $E_t$ denotes the binary Sketch response and
$M_t^{R}$ denotes the mask of body region $R$ at
frame $t$. As shown in Fig.~\ref{fig:edge_density},
TEED exhibits higher edge density, particularly in
the upper body under CL, but lower T-IoU under NT,
indicating richer local responses with greater
frame-to-frame variation.

These characteristics help explain the different
benefits of complementary representation learning.
The denser responses of TEED provide more local
variations for FOSD to reorganize, while its less
explicit global contours benefit more from Silhouette
fusion. In contrast, MUGE already preserves relatively
coherent body structures, resulting in smaller
additional gains. Overall, the effectiveness of
PolarGait is influenced by the response characteristics
of the input Sketch.

\begin{table}[!t]
    \centering
    \caption{Cross-dataset generalization results. Models are
    trained on the source dataset and evaluated on the
    target dataset.}
    \label{tab:cross_dataset_generalization}

    \scriptsize
    \setlength{\tabcolsep}{2.0pt}
    \renewcommand{\arraystretch}{1.08}

    \resizebox{0.98\columnwidth}{!}{%
    \begin{tabular}{@{}llccc|cccc@{}}
        \toprule
        \multirow{2}{*}{Input}
        & \multirow{2}{*}{Method}
        & \multicolumn{3}{c|}{CCPG $\rightarrow$ SUSTech1K}
        & \multicolumn{4}{c}{SUSTech1K $\rightarrow$ CCPG} \\
        \cmidrule(lr){3-5}
        \cmidrule(l){6-9}
        & & CL & NT & OA@R-1 & CL & UP & DN & BG \\
        \midrule

        Silh.
        & DeepGaitV2~\cite{deepgaitv2}
        & 17.6 & 14.7 & 23.0
        & \textbf{13.0} & 17.9 & 26.8 & 39.4 \\

        Parsing
        & DeepGaitV2~\cite{deepgaitv2}
        & 24.3 & 22.6 & 46.3
        & 11.0 & 17.7 & 32.8 & 60.0 \\

        Sketch (TEED)
        & DeepGaitV2~\cite{deepgaitv2}
        & 20.7 & 20.3 & 35.0
        & 8.1 & 14.6 & 22.5 & 48.9 \\

        Sketch (MUGE)
        & DeepGaitV2~\cite{deepgaitv2}
        & 22.9 & 27.0 & 44.0
        & 9.3 & 20.5 & 21.0 & 50.5 \\

        \addlinespace[1pt]
        \midrule

        RGB
        & BigGait~\cite{biggait}
        & 43.7 & \textbf{44.8} & 56.4
        & 4.5 & 11.5 & 11.9 & 45.5 \\

        RGB+Silh.
        & Gait-X~\cite{gaitx}
        & \textbf{47.9} & 43.4 & \textbf{73.4}
        & 7.2 & 15.1 & 22.2 & 56.0 \\

        \addlinespace[1pt]
        \midrule

        Sketch+Silh.+Pars.
        & SketchGait++ (TEED)~\cite{sketchgait}
        & 30.0 & 24.6 & 51.5
        & 12.3 & 22.1 & \textbf{34.2}
        & \textbf{70.3} \\

        \addlinespace[1pt]
        \midrule

        \multirow{2}{*}{Sketch+Silh.}
        & PolarGait (TEED)
        & 37.6 & 35.0 & 53.6
        & 10.8 & 20.4 & 26.2 & 57.7 \\

        & PolarGait (MUGE)
        & 41.1 & 36.4 & 56.0
        & 11.4 & \textbf{24.5} & 22.7 & 56.5 \\

        \bottomrule
    \end{tabular}%
    }
\end{table}

\subsubsection{Cross-Domain Generalization}

We further evaluate the cross-domain generalization of PolarGait between
CCPG and SUSTech1K. As shown in Table~\ref{tab:cross_dataset_generalization}, PolarGait
consistently improves the corresponding Sketch baselines in both transfer
directions. For CCPG$\rightarrow$SUSTech1K, the MUGE variant improves CL,
NT, and OA@R-1 from 22.9\%, 27.0\%, and 44.0\% to 41.1\%, 36.4\%, and
56.0\%, respectively, while the TEED variant also shows clear gains under CL
and NT. For SUSTech1K$\rightarrow$CCPG, PolarGait further improves the
Sketch baseline under several conditions, particularly CL and UP. These results
show that the complementary representations constructed by PolarGait improve
cross-dataset transferability, especially under clothing variation and nighttime
conditions.

\begin{figure}[t]
    \centering
    \includegraphics[
        width=1.0\columnwidth,
        height=0.8\textheight,
        keepaspectratio
    ]{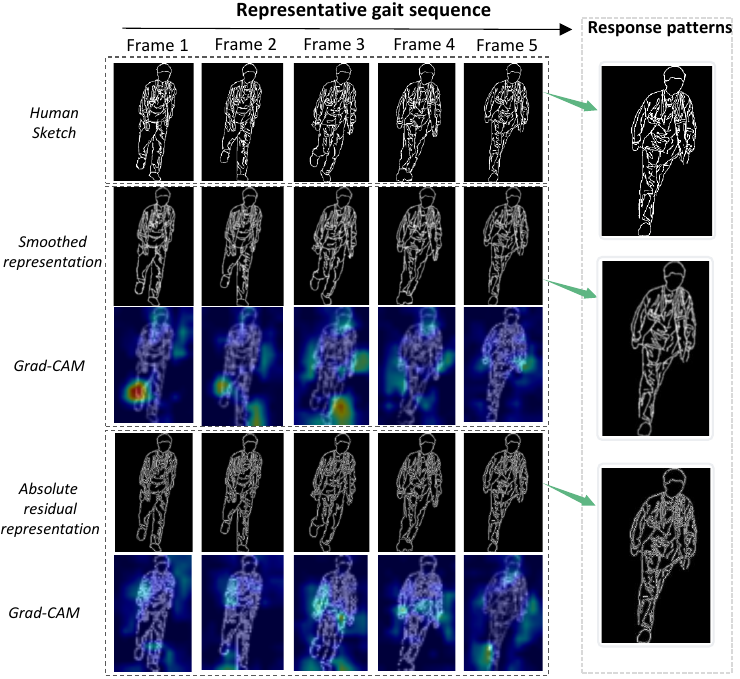}
   \caption{
Qualitative visualization of FOSD.
}
    \label{fig:gradcam}
\end{figure}

\subsection{Qualitative Visualization}

We visualize the FOSD representations and their
corresponding Grad-CAM responses across a representative
gait sequence. As shown in
Fig.~\ref{fig:gradcam}, the smoothed and
residual branches exhibit distinct spatial activation
patterns across gait frames, indicating different
discriminative emphasis despite originating from the
same Sketch. These observations provide qualitative
support for the complementary feature learning
encouraged by FOSD.

\section{Limitations and Future Directions}

PolarGait still has several limitations. First, the effectiveness of FOSD
depends on the response richness of the input Sketch, and its benefit may be
reduced when the representation is already sparse or structurally simplified.
Second, GOPT relies on predefined geometric choices, including the polar center
and the lower-body sampling ratio, which may not be optimal across different
datasets or walking conditions. In addition, the current study focuses on
Human Sketch as the primary representation for complementarity construction.
Future work could explore adaptive filtering and spatial reparameterization,
as well as whether similar principles can be extended to other gait
representations.

\section{Conclusion}

In this work, we proposed PolarGait to explore whether complementary
gait information can be actively constructed from a single representation,
rather than relying solely on heterogeneous inputs.
Using Human Sketch as the primary representation, FOSD constructs
complementary response views through Gaussian smoothing and
absolute residual extraction, while GOPT reorganizes structural
information through gait-oriented polar reparameterization.
Silhouette further provides complementary global body-shape information.
Extensive experiments demonstrate the effectiveness of these
constructed views across diverse gait recognition benchmarks.
These findings suggest that complementary gait information can
arise not only from heterogeneous representations, but also
from different response patterns and spatial organizations
within a single structured representation.
Future work will investigate adaptive view construction
and extend this principle to other gait representations.

\section*{Acknowledgements}
This work was supported by the National Natural Science Foundation
of China under Grant Nos. 62522308, 62673078, 62325302, and T2421001,
the Natural Science Foundation of Chongqing, China under Grant
No. CSTB2024NSCQ-MSX0472, and the Fundamental Research Funds for
the Central Universities under Grant Nos. 2024CDJCGJ-006,
2025CDJZKKYJH-06, and 2025CDJGF-002.

\end{document}